\def\year{2018}\relax
\documentclass[letterpaper]{article} 
\usepackage{aaai18}  
\usepackage{times}  
\usepackage{helvet}  
\usepackage{courier}  
\usepackage{url}  
\usepackage{graphicx}  
\usepackage{subcaption}
\begin{document}
%
\title{Playing Pairs with Pepper}
\author{ Abdelrahman Yaseen and Katrin Lohan\\
Heriot-Watt University\\
MACS Department\\
Edinburgh\\
}
\maketitle
\begin{abstract}
As robots become increasingly prevalent in almost all areas of society, the factors affecting humans trust in those robots becomes increasingly important. This paper is intended to investigate the factor of robot attributes, looking specifically at the relationship between anthropomorphism and human development of trust. To achieve this, an interaction game, \textit{Matching} \textit{the} \textit{Pairs}, was designed and implemented on two robots of varying levels of anthropomorphism, Pepper and Husky. Participants completed both pre- and post-test questionnaires that were compared and analyzed predominantly with the use of quantitative methods, such as paired sample t-tests. Post-test analyses suggested a positive relationship between trust and anthropomorphism with $80\%$ of participants confirming that the robots' adoption of facial features assisted in establishing trust. The results also indicated a positive relationship between interaction and trust with $90\%$ of participants confirming this for both robots post-test.
\end{abstract}
\section{Introduction}
\cite{murphy2010human} defined Human-Robot Interaction (HRI) as a field of study that involves interaction between human and robot, to address the understanding, design and evaluation of a robotic system. Recent literature has identified trust in robots as an important aspect of successful HRI and revealed that robot-related factors, including performance and attributes, have a major influence on humans’ capacity to trust a robotic system \cite{hancock2011meta}, \cite{billings2012human}, \cite{sanders2011model}. However, trust is a dynamic concept \cite{chang2010seeing} and trust in HRI is characterizes by, amongst other things,  ``the nature and characterised of the respective human team members" \cite{maurtua2017human}; as a result, there is a necessity to continually expand and develop existing research to ensure that successful HRI is sustained. For this purpose, an interaction game \textit{Matching} \textit{the} \textit{Pairs} was developed, based on Python scripts, and implemented on two contrasting robots. The game was created as a collaborative task that the human can carry out together with the robot as a team in order to explore the relationship between robot-appearance and humans' trust in robots. Anthropomorphism is a bias for attributing human characteristics to a non-human object \cite{duffy2003anthropomorphism}. Both robots were chosen to partake in the experiment for comparative purposes, that is, Pepper is a humanoid robot with facial features and the ability to communicate verbally, while, Husky is considered a machine-like, industrial-style robot only possessing the ability to communicate visually. The findings by \cite{li2010cross} suggest that a more human-like robot increases its likability and trustability, thus here we follow this suggestion. 
\section{Experiment}
\begin{figure}
        \includegraphics[width=0.5\textwidth]{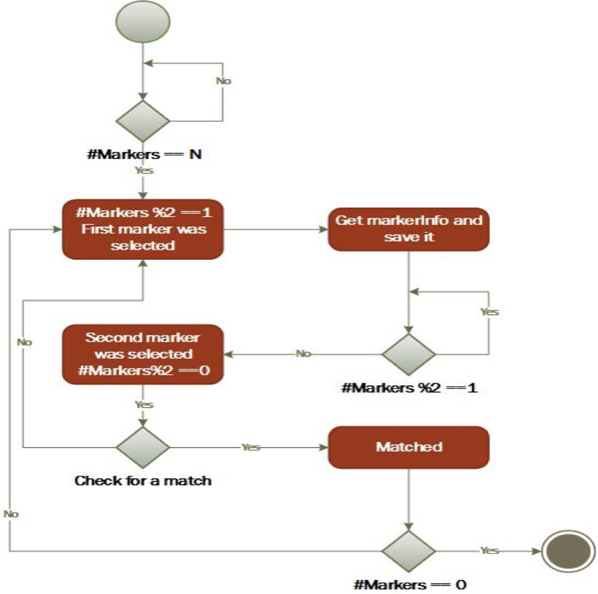}
        \caption{Activity diagram of the Finite State Machine (FSM).}
        \label{pic1}
\end{figure}

\textit{Matching the Pairs} was implemented using a Finite State Machine (FSM). The algorithm and the rules for the interaction game are described in the activity diagram (see Figure \ref{pic1}). The game consists of twelve mini-blocks; the goal is to find all six matching pairs. The participant will choose a marker (see Figure \ref{pic1}). The marker information will be saved including the symbol which it represents; to enable the robot to track the markers. Then the participant has the following two options regarding the selection of the second marker:
\begin{itemize}
\item	Ask the robot for help in choosing the other matching pair by saying "Help me". In this case, Pepper/Husky will recognize that the participant wants help via the use of the speech recognition engine and respond accordingly. Pepper will respond verbally, while Husky will respond visually i.e. display the results on the screen.
\item Choose the marker without asking the robot for help. 
\end{itemize}
Before participants started and after they finished the experiment, they were asked to fill in an adaptation of the Godspeed questionnaire \cite{bartneck2009measurement}, containing additional questions related to the experiment. Those additional questions will be our primary measure of trust and the Godspeed is used to compare how participants felt about the robot before and after exposure. The questionnaires -- pre and post -- were compared using a paired sample t-test. Once the game began, Pepper/Husky indicated the rules of the game, and provided guidance to the participants throughout the game. A low-error condition was incorporated on the robots with the occurrence of mistakes $3\%$ of the time throughout the game. The game finished once all pairs were found.

\begin{figure}
    \centering
    \begin{subfigure}[b]{0.27\textwidth}
        \includegraphics[width=\textwidth]{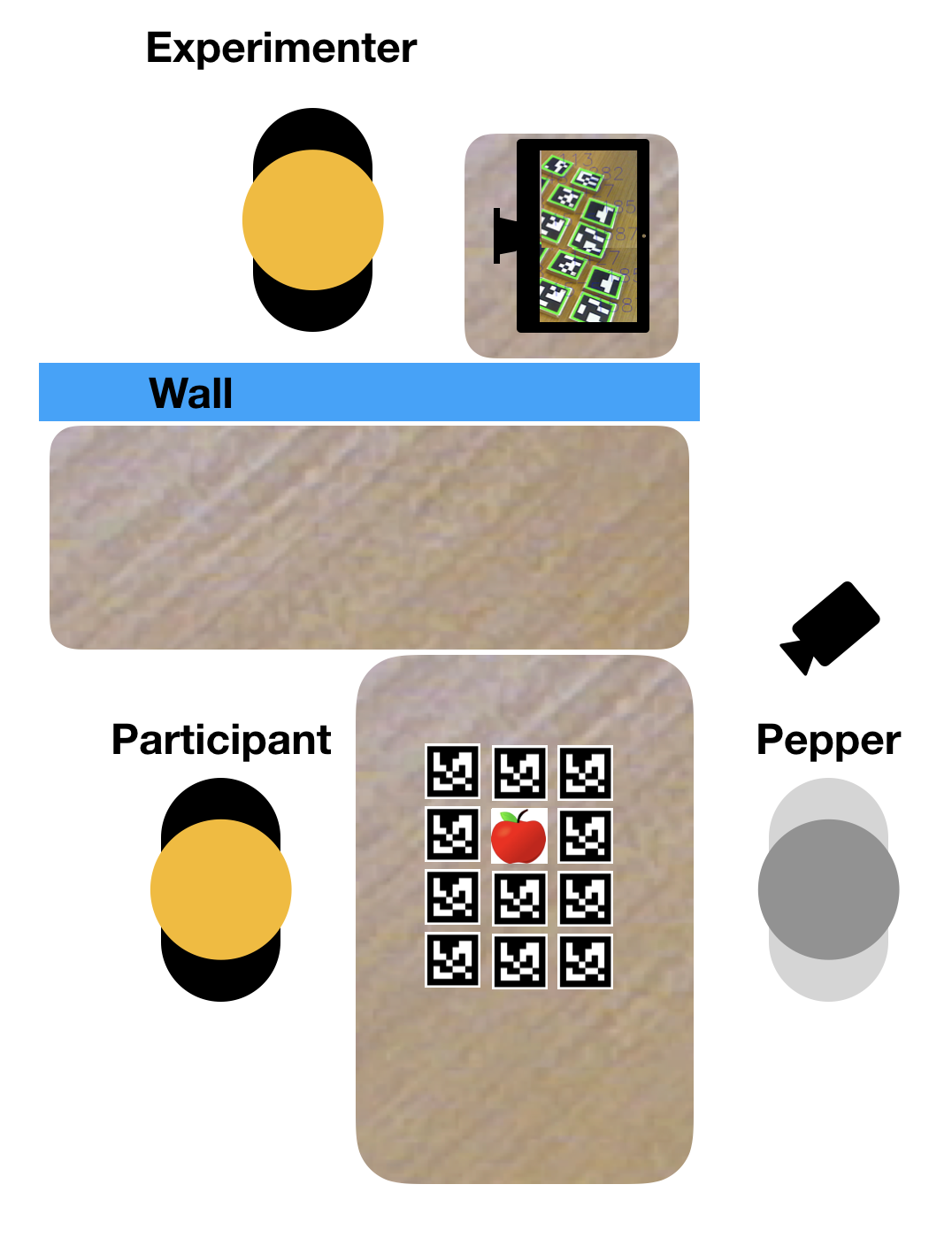}
        \caption{The setup.}
        \label{pic2}
    \end{subfigure}
    \begin{subfigure}[b]{0.12\textwidth}
        \includegraphics[width=\textwidth]{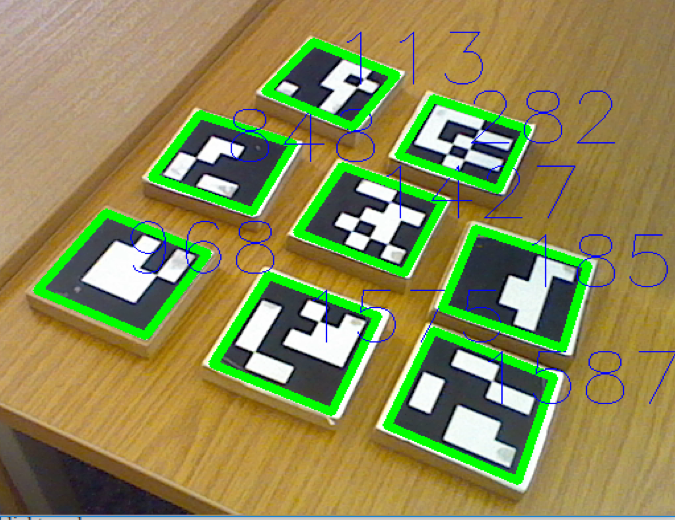}
        \caption{Fiducial marker used with the AR-marker-detection library}
        \label{pic3}
    \end{subfigure}
    \caption{Experimental setup and system}\label{fig:animals}
\vspace{-0.5cm}
\end{figure}

Twenty participants from Heriot-Watt University were randomly assigned to play the interaction game with either robot. The sample consisted of seventeen students and three researchers, fourteen males and six females. The mean age of the participants was 24.7 years. An information sheet was handed to each participant with information regarding the robot they would be interacting with along with an image of that robot. A written and approved consent form was obtained from each participant prior to the experiment. 
The setup can be seen in \ref{pic2}. The blocks used to play pairs are fitted on the side with a marker (see Figure \ref{pic3}) and on the other side with images of fruits.
A camera was used to recognize these markers and capture the game to display it for the experimenter. The role of the experimenter was to monitor the execution of the game via a screen; the experimenter was not visible to the participant. 
\section{Results and Discussion  }
Participants perceived Pepper to be significantly more trustworthy after interaction: pre-test (M = 2.8, SD = 0.92) post-test (M = 3.9, SD = 0.99), t(9) = -2.4, p = 0.04. While Husky’s perceived trustworthiness did not increase significantly following exposure with: pre-test (M = 3.1, SD = 1.37), post-test (M = 3.9, SD = 0.994), t(9) = -1.45, p = 0.182. Before interaction, participants perceived Pepper as: interactive, lively, social, friendly, and relaxed. Participants perceived Pepper to be significantly more human-like after interaction with: pre-test (M = 2, SD = 0.82) post-test (M = 3.2, SD = 1.14), t(9) = -2.34, p = 0.044. In comparison, before exposure participants perceived Husky as: inert, stagnant, apathetic, and anti-social. Husky was perceived as machine-like before and after interaction with: pre-test (M = 1.8, SD = 1.03) post-test (M = 2.2, SD = 1.23), t(9)= -0.71, p = 0.49. The use of facial features was confirmed as having a facilitating effect on developing trust in Pepper with $80\%$ of participants confirming this in both tests. The use of both verbal and visual communication by Pepper and Husky respectively increased levels of trust for $90\%$ of participants; the anthropomorphic trait of a ‘voice’ did not contribute to trust any more than the use of visual aids. Furthermore, $90\%$ of participants believed that interaction assisted in increasing their trust in the robots following exposure to Pepper and Husky.
\vspace{-0.5cm}
\section{Conclusion}
The aim of this paper was to investigate the relationship between human trust and robot appearance. After running the experiment and analyzing the results, we surmise that anthropomorphising robots’ physical appearance and increasing the occurrence of human-robot interaction are effective means of increasing human trust levels in their robot counterparts. A positive relationship was established between robot anthropomorphism and trust; the more anthropomorphic the robot, the more it can be trusted. The adoption of facial features on the robot is a technique used for anthropomorphising objects towards human-likeness and encouraged participants to trust Pepper more than Husky. It is worth noting that the outcome of interacting with a robot is highly dependent on the human involved in the interaction, that is, some participants were utterly reliant on the robot, others used it as a consultant, whilst others did not depend on the robot at all. This individuality and context specifically emphasizes the demand for further research to be undertaken to continually develop our understanding of HRI.
\section{Acknowledgements}
The authors would like to acknowledge the support of the EPSRC IAA 455791 along with ORCA Hub~EPSRC (EP/R026173/1, 2017-2021) and consortium partners. We would also like to achknowledge our colaborators from the University of Bielefeld, Prof. Dr. Friederike Eyssel and Jasmin Bernotat for the discussions on this project. 
\vspace{-0.2cm}
\bibliographystyle{aaai}
\bibliography{MtPGame} 
\end{document}